\documentclass[9pt,conference]{IEEEtran}
\IEEEoverridecommandlockouts

\usepackage{cite}
\usepackage{amsmath,amssymb,amsfonts}
\usepackage{algorithmic}
\usepackage{graphicx}
\usepackage{textcomp}
\usepackage{xcolor}
\usepackage{tabularray}
\usepackage{booktabs}
\usepackage{multirow}
\usepackage{makecell}
\usepackage{bbding}
\usepackage{pifont}
\usepackage{wasysym}
\usepackage{utfsym}
\usepackage{fontawesome}

\def\BibTeX{{\rm B\kern-.05em{\sc i\kern-.025em b}\kern-.08em
    T\kern-.1667em\lower.7ex\hbox{E}\kern-.125emX}}

\begin{document}
\title{``I've Heard of You!": Generate Spoken Named Entity Recognition Data for Unseen Entities}



\author{Jiawei Yu$^{1*}$, Xiang Geng$^{2*}$, Yuang Li$^{3}$, Mengxin Ren$^{3}$, Wei Tang$^{3}$, Jiahuan Li$^{2}$, Zhibin Lan$^{1}$,\\ Min Zhang$^{3}$, Hao Yang$^{3}$, Shujian Huang$^{2\dagger}$, Jinsong Su$^{1\dagger}$\thanks{* denotes equal contribution to this work.}\\ 
\thanks{$\dagger$ Corresponding author}
$^{1}$School of Informatics, Xiamen University, China\\
$^{2}$National Key Laboratory for Novel Software Technology, Nanjing University, China
\\
$^{3}$Huawei Translation Services Center, China\\
\small{yujiawei@stu.xmu.edu.cn}, \small{gx@smail.nju.edu.cn}, \small{jssu@xmu.edu.cn} \\
\small{\{liyuang3,zhangmin186,yanghao30\}}@huawei.com}

\maketitle

\begin{abstract}
Spoken named entity recognition (NER) aims to identify named entities from speech, playing an important role in speech processing.
New named entities appear every day, however, annotating their Spoken NER data is costly. 
In this paper, we demonstrate that existing Spoken NER systems perform poorly when dealing with previously unseen named entities.
To tackle this challenge, we propose a method for generating Spoken NER data based on a named entity dictionary (NED) to reduce costs.
Specifically, we first use a large language model (LLM) to generate sentences from the sampled named entities and then use a text-to-speech (TTS) system to generate the speech.
Furthermore, we introduce a noise metric to filter out noisy data.
To evaluate our approach, we release a novel Spoken NER benchmark along with a corresponding NED containing 8,853 entities.
Experiment results show that our method achieves state-of-the-art (SOTA) performance in the in-domain, zero-shot domain adaptation, and fully zero-shot settings. Our data will be available at https://github.com/DeepLearnXMU/HeardU.

\end{abstract}


\section{Introduction}
Spoken named entity recognition (NER) is the task of identifying and categorizing named entities from speech into pre-defined categories, such as {person} (PER), {location} (LOC), and {organization} (ORG) \cite{whereNER}. It is typically performed using either a pipeline system ~\cite{sudoh2006incorporating,raymond2013robust,jannet2015evaluate,Ruan2020} or an end-to-end (E2E) system \cite{e2e_SNER,yadav2020end,AISHELL-NER,arora2022token}. Spoken NER is crucial for grasping the meaning of speech and therefore has been widely applied, including protecting entity privacy~\cite{Audio_De-identification} and correcting entity errors in automatic speech recognition (ASR)~\cite{ASR_Entity_Correction}.

However, Spoken NER poses greater challenges than text NER.
Speech varies greatly in pronunciation, accents, and dialects, significantly expanding the input space beyond that of text.
Additionally, annotating Spoken NER data is more costly than annotating standard text NER data.
Consequently, the availability of open-source Spoken NER datasets is limited, focusing on specific domains and languages~\cite{msner}.
Every day, numerous new entities are named, such as newly released products. Can existing Spoken NER methods recognize these unseen entities by using Spoken NER data from other domains?

\begin{figure}[t]
\centering
  \includegraphics[width=0.8\linewidth]{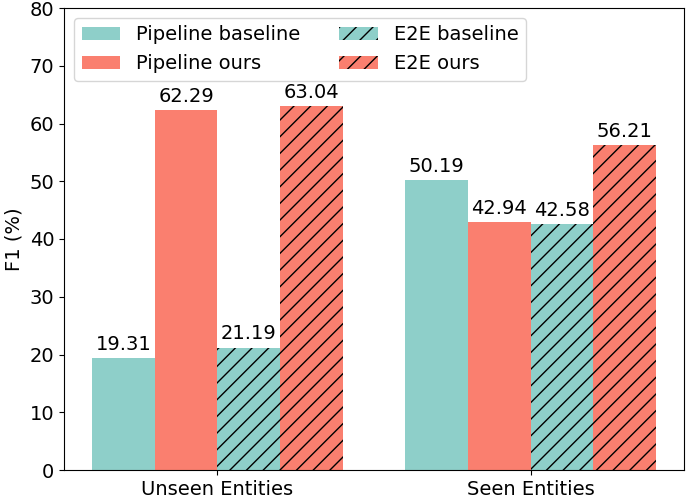}
  \vspace{-0.3cm}
  \caption{
  Both pipeline and E2E Spoken NER methods perform worse on unseen entities compared to seen entities.
  Our method works well, with greater improvement observed for unseen entities.}
  \label{fig:f1 unseen and seen}
  \vspace{-0.6cm}
\end{figure}

Unfortunately, as illustrated in Figure~\ref{fig:f1 unseen and seen}, both pipeline and E2E Spoken NER methods show much poorer performance on unseen entities compared to seen entities.\footnote{Unseen entities are defined as those that appear in the target domain test set, but not in the general-domain training set.} 
Previous work, Un-Sp\cite{External_SNER_Data} attempts to generate Spoken NER data by distilling different Spoken NER models.
However, this method highly relies on in-domain speech which is hard to acquire, and the noise of generated data is hard to measure.
In this paper, we propose a novel framework named HeardU, which generates Spoken NER data based on the named entity dictionary (NED).
Building the entity dictionary is much easier than acquiring Spoken NER data or in-domain speech.
Moreover, the entity dictionary is off-the-shelf in many downstream applications, including terminology databases in machine translation~\cite{constrained_decoding,constrained_decoding2} and name sets in dialogue generation~\cite{li2016diversity}.

\begin{figure}[!t]
\includegraphics[width=\linewidth]{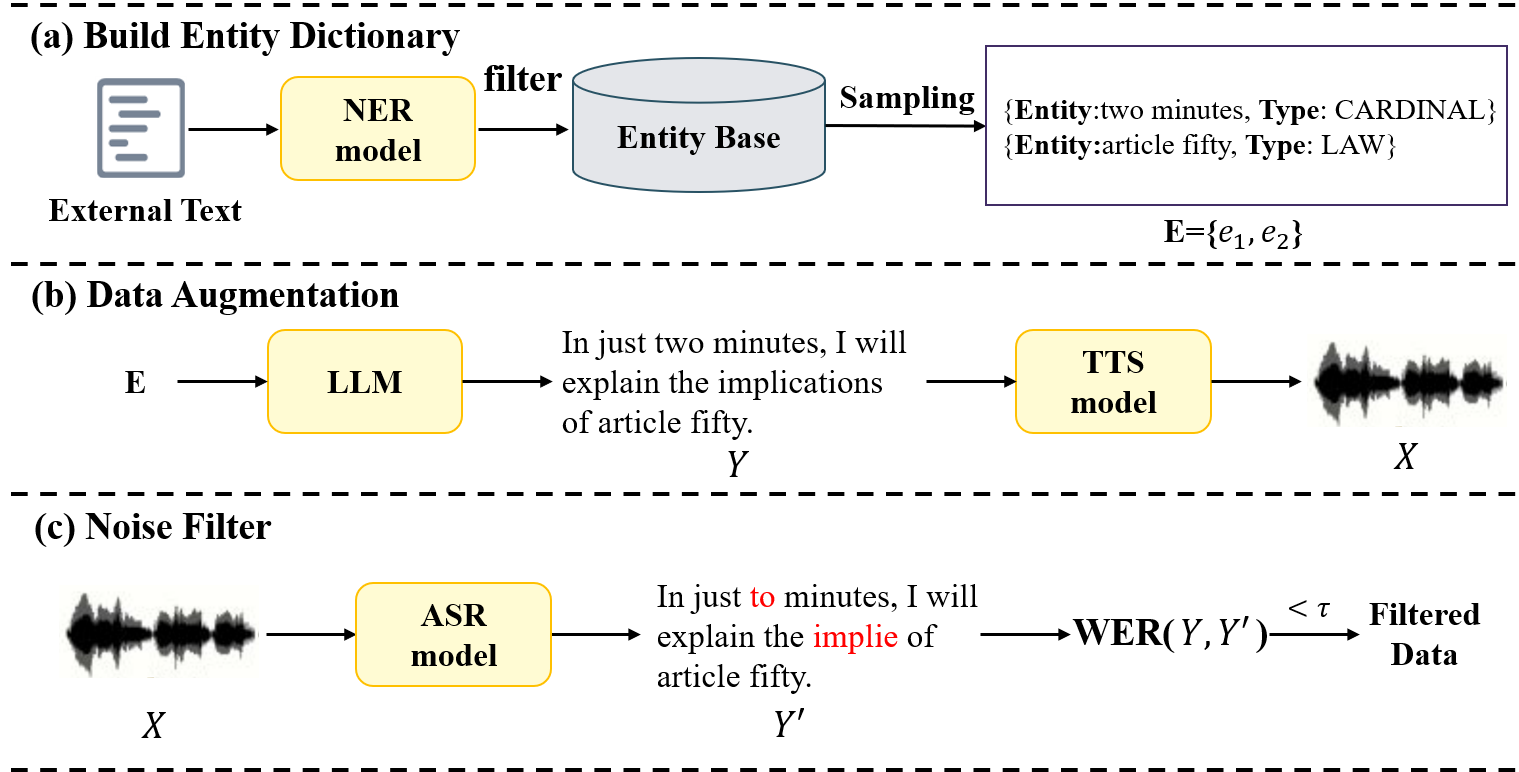}
    \vspace{-0.6cm}
  \caption {The HeardU framework.}
  \label{fig: framework}
  \vspace{-0.5cm}
\end{figure}

The HeardU pipeline is presented in the Figure \ref{fig: framework}.
To construct an NED effectively, we employ a general-domain NER model to recognize entities within the in-domain documents and ask the large language model (LLM) or human annotators to refine the recognized entities.
Next, we sample multiple entities from the NED and instruct the LLM to generate coherent contextual information based on these selected entities and their respective types.
Finally, we feed the generated text into a general-domain text-to-speech (TTS) model to synthesize the corresponding speech output.
We analyze the noise source of the generated Spoken NER data and propose a noise metric to identify low-quality data.

For evaluation, we introduce a new benchmark for Spoken NER in Chinese, named ST-CMDS-NER, along with its associated human-refined NED.
Experiment results show that HeardU achieves state-of-the-art (SOTA) results across different settings.
Using the LLM-refined NED in English, HeardU achieves F1 score increases of 2.76\%, 19.76\%, and 18.33\% in the in-domain, zero-shot domain adaptation, and fully zero-shot settings, respectively.
Using the human-refined NED in Chinese, HeardU achieves F1 score increases of 32.98\% and 9.66\% in the zero-shot domain adaptation and fully zero-shot settings, respectively.


Overall, our contributions are summarized as follows:
\begin{itemize}
    \item We show that existing Spoken NER methods fail to handle unseen entities and highlight that building the NED is more feasible than acquiring Spoken NER data.
    \item We release a new Spoken NER benchmark in Chinese and its corresponding NED for zero-shot domain adaptation and fully zero-shot settings.
    \item We propose the HeardU framework to generate Spoken NER data by using the NED, LLM, and TTS model, and to identify and filter out low-quality data. 
\end{itemize}


\section{Methodology}

\subsection{Task Formulation}

In Spoken NER, an instance is generally represented as a triplet $\{X, Y, Z\}$. Here, $X$ refers to the spoken utterance, $Y=\{y_1, y_2,\dots, y_n \}$ represents its transcription with $n$ tokens, and $Z=\{z_1, z_2,\dots, z_n \}$ denotes the associated BIO tags. 
For BIO tags~\cite{BIO}, B-$t$ and I-$t$ indicate the beginning and inner tokens of an entity categorized as type $t$, where $t$ belongs to a pre-defined set of entity types $T=\{t_1, t_2,\dots, t_m \}$.
O represents tokens that do not belong to any named entity. 

\subsection{Named Entity Dictionary}
As discussed before, many downstream tasks already have an NED. This section shows that constructing an NED from scratch is much more efficient and cost-effective than creating Spoken NER datasets, making HeardU adaptable to various scenarios.
Inspired by AISHELL-NER \cite{AISHELL-NER}, we use a text NER model for automatic NED construction. Specifically, we use general-domain NER data to train a text NER model. Then, we use the NER model to get coarse-grained named entities on in-domain documents. The results are very noisy due to the wrong boundary or hallucination of the entity. Therefore, we merge similar entities and ask the LLM or human annotators to refine them. We provide instructions for LLM refinement in Table \ref{prompt_en}. Please note that refining the NED is more efficient than labeling named entities sentence by sentence.
In practice, we use the LLM-refined NED and human-refined NED in English and Chinese experiments, 
respectively.


\subsection{Data Augmentation}
To generate Spoken NER data, we first sample entities from the NED. In human speech, the distribution of entities is usually sparse and long-tail. Some sentences may contain no entities, and some entities may only appear once in the whole dataset. To achieve better performance on long-tail entities, we try to generate a dataset with balanced entities. As such, we uniformly and randomly sample one or two entities for one sentence (the most common cases) from the NED. After that, we can get the sampled entities $E=\{e_1 \}$ or $E=\{e_1, e_2 \}$.

\begin{table}[t]
\centering
\vspace{-0.4cm}
\caption{Instructions for data filtering in Figure \ref{fig: framework} (a) and data generation in Figure \ref{fig: framework} (b)}
\resizebox{\columnwidth}{!}{
\begin{tabular}{c|l} 
\toprule
\multirow{2}{*}{\begin{tabular}[c]{@{}c@{}}Refine \\entity\end{tabular}} & \begin{tabular}[c]{@{}l@{}}I will write you an entity and its type, you need to judge \\If it is an entity and the type is correct, print YES or NO.\end{tabular}                                                                                                   \\
                                                                         & \begin{tabular}[c]{@{}l@{}}\#\#\#User: My entity is `$\{e_1 \}$', the type is `$\{t_1 \}$'\\\#\#\#\#Response: ...\end{tabular}                                                                                                                                                  \\ 
\hline
\multirow{3}{*}{\begin{tabular}[c]{@{}c@{}}Generate\\text\end{tabular}}  & \begin{tabular}[c]{@{}l@{}}I want you to act as a speaker in $\{domain\}$. I will write you\\entities and their type,~you need to output a sentence\\containing these entities.~The resulting sentence should be\\more than 20 words and less than 100 words.\end{tabular}  \\
                                                                         & \#\#\#User: My entities are `$\{e_1, e_2 \}$', the types are `$\{t_1,t_2\}$'                                                                                                                                                                                                          \\
                                                                         & \#\#\#\#Response: ...                                                                                                                                                                                                                                                    \\
\bottomrule
\end{tabular}
}
\vspace{-0.7cm}
\label{prompt_en}
\end{table}

Then we instruct the LLM to act as the speaker in the target domain and generate a sentence containing the sampled entities based on their types.
In preliminary studies, we found that TTS models perform badly on very short sentences. Besides, human speakers seldom speak very long sentences. Therefore, we ask the LLM to generate a sentence with 20-100 words. The instructions are provided in Table \ref{prompt_en}.
We take the generated sentence as the transcription $Y$ and we can get the BIO tags $Z$ and E2E target $\hat{Y}$ easily by lexical matching.

Finally, we input the generated sentence into a TTS model for generating its speech $X$. To generate speech with diverse audio features, we randomly choose a speaker of the TTS model and perturb the speaking speed.
We further add background noise and simple sound effects to the speech to improve the robustness of Spoken NER models.
\subsection{Noise Filtering}
We start with analyzing the noise sources in the generated Spoken NER data: the BIO tags $Z$, the transcription $Y$, and the speech $X$.
The generated BIO tags may often go wrong in the following cases: the sampled entity is noisy due to its incomplete or redundant text or wrong type; the LLM generates text without following the required entity type \footnote{for example, generating ``Apple" with the type ``fruit" instead of ``company".}.
Due to the satisfying NED and powerful LLM, these errors seldom happen.
To confirm that, we manually label 50 random instances for the LLM-refined NED and human-refined NED separately.
The two NEDs used in our experiments produce accurate labels, achieving 86\% and 96\% accuracy, respectively.

The generated text for data augmentation should be fluent and similar to the target distribution; otherwise, it will significantly degrade performance despite the accurate labels.
The LLM, trained on extensive and diverse data, is supposed to generate fluent and reasonable text even in specific domains. 
To examine the assumption, we evaluate the transcription using the language model (LM) in the target domain. 
The average perplexity of synthetic text is close to real text in the target domain (187.67 vs. 165.55).

Most of the noise is introduced by the TTS model.
Usually, the TTS model is evaluated by humans, and it is difficult to automatically evaluate the quality of generated speech \cite{tan2024naturalspeech}.
Besides, the evaluation target differs in this task, where we aim to generate speech that an ASR or entity-aware ASR model can recognize.
Therefore, we introduce a method similar to round-trip translation for machine translation quality estimation \cite{moonrevisiting}.
Specifically, we first use an ASR model to generate the transcription $Y'$ for the speech $X$, and then we calculate the similarity between $Y$ and $Y'$ using Word Error Rate (WER). 
With this noise metric, we can filter out the noisy data by simply setting a threshold $\tau$.

\section{New Benchmark}

To the best of our knowledge, AISHELL-NER \cite{AISHELL-NER} is the only annotated NER dataset derived from Chinese speech. 
Based on a public ASR dataset ST-CMDS \cite{st-cmds}, we introduce a new Spoken NER benchmark ST-CMDS-NER in Chinese.
The speech in ST-CMDS mainly focuses on two topics: usual online voice chats and intelligent voice control statements. For construction, we first sample 3,000 sentences from ST-CMDS. Following AISHELL-NER, we annotate three types of named entities for each sentence: PER, LOC, and ORG. As for the construction of our human-annotated NED, we sample other 40,000 sentences from the rest of ST-CMDS and use an NER model trained on AISHELL-NER to predict the named entities.
Next, we manually refine these named entities by deleting the non-entity words and correcting the boundary of the wrong entity words. Finally, we get 8,853 entities to build our NED. Besides, the entity overlap between the Spoken NER test set and the NED is 98.70\%, which indicates the NED we annotated is of high quality.
We also hire eight professional annotators for data labeling work and an expert to sample and check the annotations for quality control. Table \ref{ST-CMD-static} provides detailed information on this benchmark. To estimate the performance of manual annotations, following \cite{slue}, we select 30\% of the data in the dataset for a second annotation. The second pass achieves an F1 score of 92.1\% when evaluated against the first pass. This indicates a high degree of consistency in the dataset we annotated.
We also measure the efficiency of labeling various types of data (instance per hour) in Table \ref{time-cost}. Building an NED is approximately 24 times faster than annotating Spoken NER data, indicating that constructing an NED is much more efficient and cost-effective than creating Spoken NER datasets.

\begin{table}[t!]
\centering
\vspace{-0.3cm}
\caption{Statistics of the AISHELL-NER dataset and our annotated ST-CMDS-NER and corresponding NED.}
\vspace{-0.2cm}
\tabcolsep=0.07cm
\resizebox{\linewidth}{!}{
\begin{tabular}{lccccc} 
\hline
\textbf{\textbf{\textbf{\textbf{Dataset}}}} & \textbf{~~\#Sentence} & \textbf{~~w/NE} & \textbf{~~\#PER} & \textbf{~~\#LOC} & \textbf{~~\#ORG}  \\ 
\hline
AISHELL-NER                                 & 120,098               & 40,839          & 15,842           & 20,693           & 21,455            \\
ST-CMDS-NER                                 & 3,000                 & 1,004           & 700              & 369              & 202               \\
ST-CMDS-NED                                & N/A                   & N/A             & 5,741            & 1,737            & 1,372             \\
\hline
\end{tabular}
}
\label{ST-CMD-static}
\vspace{-0.4cm}
\end{table}

\begin{table}[t!]
\centering
\caption{Time cost of different data annotation.}
\vspace{-0.2cm}
\label{time-cost}
\renewcommand\arraystretch{1.2}
\begin{tabular}{lllll} 
\toprule
                                                                             & NED                      & Text NER                 & ASR                     & Spoken NER               \\ 
\hline
\begin{tabular}[c]{@{}l@{}} Num of Entities / Hour\end{tabular} & \multicolumn{1}{c}{3660} & \multicolumn{1}{c}{840} & \multicolumn{1}{c}{180} & \multicolumn{1}{c}{148}  \\
\bottomrule
\end{tabular}
\vspace{-0.3cm}
\end{table}

\section{Experiments}
\subsection{Settings}
We conduct our experiments in three settings: the in-domain setting, where the in-domain training data is available; the zero-shot domain adaptation setting, where only the source general-domain training data is available; and the fully zero-shot setting, where human-labeled training data is unavailable in both target and general domains. Note that in all three settings, we test our models in the target in-domain test set. Table \ref{settings} shows all settings and available datasets for training our model.

\begin{table}[t]
\centering
\caption{Experimental settings and available datasets for training our model.}
\renewcommand\arraystretch{1.25}
\resizebox{\linewidth}{!}{
\begin{tabular}{lccc} 
\toprule
Settings          & \begin{tabular}[c]{@{}c@{}}General-domain \\data (real)\end{tabular} & \begin{tabular}[c]{@{}c@{}}In-domain \\data (real)\end{tabular} & \begin{tabular}[c]{@{}c@{}}In-domain \\data (pseudo)\end{tabular}  \\ 
\hline
In-domain                       & {\ding{55}}     &      {\checkmark}                        &      {\checkmark}                                   \\
Zero-shot Domain adaptation  & {\checkmark}     &     {\ding{55}}                          &                  {\checkmark}                      \\
Fully zero-shot             & {\ding{55}}     &                      {\ding{55}}        &         {\checkmark}                                \\
\bottomrule
\end{tabular}}
\label{settings}
\end{table}

\paragraph{Data}
In English (En) experiments, we use SLUE-Voxpopuli \cite{slue} test set in the European Parliament domain as our test set, which is about 5 hours.
In the in-domain setting, we use SLUE-VoxPopuli training set for training, which is about 15 hours.
In the domain adaptation setting, we use SLURP \cite{SLURP} with nearly 100 hours of speech as our general-domain dataset.
In Chinese (Zh) experiments, we use our annotated benchmark ST-CMDS-NER in the intelligent voice control domain as the test set.
In the domain adaptation setting, we use AISHELL-NER as our general-domain dataset, which is annotated based on AISHELL-1 \cite{aishell-1} that contains 170 hours of speech.
Note that in this setting, we use the human-refined NED to generate Spoken NER data.

\begin{table}[!t]
\caption{The performance of different Spoken NER methods in the \textbf{in-domain, zero-shot domain adaptation}, and \textbf{fully zero-shot} settings.}
\centering
\renewcommand\arraystretch{1.1}
\tabcolsep=0.08cm
\resizebox{\linewidth}{!}{
\begin{tabular}{l|lccccc} 
\toprule
\multicolumn{1}{l}{Lang} & Method                                                                                                                                                 & WER$\downarrow$         & Precision$\uparrow$     & Recall$\uparrow$        & F1$\uparrow$            & label-F1$\uparrow$  \\ 
\hline
\multirow{14}{*}{En}     & \multicolumn{6}{l}{\textbf{\textit{Pipeline}}}                                                                                                                                                                                                                                       \\
                         & GD\cite{slue}                                                                                                                                                     & 47.48                   & 24.04                   & 8.62                    & 12.69                   & 26.65               \\
                         & \textbf{HeardU}                                                                                                                                        & \underline{18.13}           & 31.16                   & \underline{35.08}           & 33.01                   & \underline{48.26}       \\
                         & GD+\textbf{HeardU}                                                                                                                                     & 20.20                   & \underline{31.45}           & 34.37                   & \underline{33.06}           & 48.14               \\
                         & ID\cite{slue}                                                                                                                                                     & 12.06                   & 69.32                   & 70.07                   & 69.69                   & 80.50               \\
                         & ID+Un-Sp\cite{External_SNER_Data}                                                                                                                                               & 11.52                   & \textbf{72.54}          & 68.43                   & 70.43                   & 79.82               \\
                         & ID\textbf{\textbf{+HeardU}}                                                                                                                            & \textbf{11.27}          & 71.14                   & \textbf{\textbf{75.17}} & \textbf{\textbf{73.10}} & \textbf{82.62}      \\ 
\cline{2-7}
                         & \multicolumn{6}{l}{\textbf{\textit{E2E}}}                                                                                                                                                                                                                                            \\
                         & GD\cite{slue}                                                                                                                                                     & 52.68                   & 24.14                   & 7.87                    & 11.86                   & 21.47               \\
                         & \textbf{\textbf{HeardU}}                                                                                                                               & 31.41                   & 36.57                   & \underline{25.70}           & 30.19                   & 41.09               \\
                         & GD+\textbf{\textbf{HeardU}}                                                                                                                            & \underline{31.02}           & \underline{42.31}           & 25.24                   & \underline{31.62}           & \underline{41.97}       \\
                         & ID\cite{slue}                                                                                                                                                     & 17.27                   & 70.67                   & 59.36                   & 64.52                   & 73.73               \\
                         & ID+Un-Sp\cite{External_SNER_Data}                                                                                                                                               & \textbf{\textbf{14.13}} & 73.81                   & 68.43                   & 71.02                   & 78.43               \\
                         & ID\textbf{\textbf{\textbf{\textbf{+HeardU}}}}                                                                                                          & 14.45                   & \textbf{\textbf{77.53}} & \textbf{\textbf{70.37}} & \textbf{\textbf{73.78}} & \textbf{79.12}      \\ 
\hline
\multirow{8}{*}{Zh}      & \multicolumn{6}{l}{\textbf{\textit{Pipeline}}}                                                                                                                                                                                                                                       \\
                         & GD\cite{slue}                                                                                                                                                     & 23.73                   & 33.63                   & 26.99                   & 29.94                   & 63.64               \\
                         & \textbf{\textbf{\textbf{\textbf{\textbf{\textbf{\textbf{\textbf{\textbf{\textbf{\textbf{\textbf{\textbf{\textbf{\textbf{\textbf{HeardU}}}}}}}}}}}}}}}} & 30.00                   & 23.66                   & \textbf{63.10}          & 34.41                   & 46.73               \\
                         & GD+\textbf{\textbf{\textbf{\textbf{\textbf{\textbf{\textbf{\textbf{HeardU}}}}}}}}                                                                      & \textbf{22.32}          & \textbf{49.34}          & 62.23                   & \textbf{55.05}          & \textbf{74.67}      \\ 
\cline{2-7}
                         & \multicolumn{6}{l}{\textbf{\textit{E2E}}}                                                                                                                                                                                                                                            \\
                         & GD\cite{slue}                                                                                                                                                     & 27.76                   & 41.05                   & 21.09                   & 27.86                   & 53.64               \\
                         & \textbf{\textbf{\textbf{\textbf{\textbf{\textbf{\textbf{\textbf{\textbf{\textbf{\textbf{\textbf{\textbf{\textbf{\textbf{\textbf{HeardU}}}}}}}}}}}}}}}} & 43.59                   & 27.92                   & \textbf{57.20}          & 37.52                   & 49.45               \\
                         & GD+\textbf{\textbf{\textbf{\textbf{\textbf{\textbf{\textbf{\textbf{HeardU}}}}}}}}                                                                      & \textbf{18.30}          & \textbf{72.22}          & 52.56                   & \textbf{\textbf{60.84}} & \textbf{69.13}      \\
\bottomrule
\end{tabular}
}
\label{main-exp}
\vspace{-0.6cm}
\end{table}

\paragraph{Implementation}
We use gpt-3.5-turbo-0125 as our LLM for text generation and entity refinement in section II. We use Bark-TTS\footnote{https://github.com/suno-ai/bark} and Chat-TTS\footnote{https://github.com/2noise/ChatTTS} as our TTS toolkits for En and Zh speech generation, respectively. We set the noise filter threshold $\tau$ to 0.5 for En experiments and 0.3 for Zh experiments.
The wav2vec 2.0 base \cite{wav2vec2} and DeBERTa base \cite{deberta} are used as the unsupervised pre-trained models.
We use the fairseq library \cite{fairseq} to fine-tune wav2vec 2.0 models for the E2E NER and ASR tasks with 80k updates on 100 hours of pseudo-labeled data. We use HuggingFace’s transformers toolkit \cite{huggingface} to train the text NER model on pseudo-labels. Our E2E system uses entity-aware ASR. It adds special tokens to the ASR vocabulary list to identify named entities, $i.e.$, $[$ $]$ for PER, $( )$ for LOC, and $< >$ for ORG. Following \cite{slue}, we use the LM for decoding with beam size 500, LM weight 2, and word insertion penalty -1. All LMs are trained on the corresponding training set in each setting.

\paragraph{Evaluation metrics}
We use WER (\%) as the ASR evaluation metric. 
The WER is calculated based on word- and character-level for En and Zh, respectively.
We use F1 (\%) and Label-F1 (\%) as the NER evaluation metrics. The F1 score is the harmonic mean of precision and recall, and Label-F1 considers only the tag predictions.


\paragraph{Baselines}
(1) Vanilla Pipeline and E2E methods \cite{slue}. In our in-domain setting, the baseline wav2vec 2.0 models are trained using in-domain (ID) human-labeled data. Zero-shot domain adaption and fully zero-shot settings share the same data, where the models are trained only using general-domain (GD) human-labeled data. (2) Un-Sp \cite{External_SNER_Data}. It uses external 100 hours of in-domain speech to enhance the model with distillation; thus, we only compare it in the in-domain setting.

\subsection{Main Results}

As shown in Table \ref{main-exp}, our HeardU outperforms the baselines in all settings. 
Specifically, HeardU achieves a 32.98\% increase in F1 score in the Zh zero-shot domain adaptation setting (GD+HeardU) by using a human-refined high-quality NED. Besides, with an automatically constructed NED, HeardU shows substantial improvements in both En zero-shot domain adaptation (GD+HeardU) (+19.76\% F1 scores) and in-domain settings (ID+HeardU) (+2.76\% F1 scores). Moreover, our method achieves the SOTA performance in the Label-F1 metric.


In the Zh fully zero-shot setting, the HeardU line shows a very high WER, which indicates that the test speech cannot be accurately recognized, thus greatly affecting the final F1 score. This is attributed to the limited capabilities of TTS models and the resulting poor-quality synthetic speech. 
However, HeardU still outperforms the vanilla domain adaptation method because it identifies more target entities in the test set and achieves higher recall scores.


\section{Analysis}


\begin{table}[t!]
\vspace{-0.3cm}
\centering
\caption{The cases generated by our method.}
\resizebox{\columnwidth}{!}{
\begin{tabular}{l} 
\toprule
\begin{tabular}[c]{@{}l@{}}Entity: \textbf{Salva Kiir}, Type: \textbf{PER}\\User: \{Generate prompt illustrated in Table \ref{prompt_en}\} \\\textbf{}Response:~\textbf{Salva Kiir} is a prominent political figure in South \\Sudan, serving as the country's president.~ ~\end{tabular}                                                                                                                                                                              \\ 
\hline
\begin{tabular}[c]{@{}l@{}}Entity:~\textbf{\textbf{Sakorafa}}, Type: \textbf{PER}\\User: Do you understand entity \textbf{Sakorafa}? Give its meaning.\\Response: I'm sorry, but ``\textbf{Sakorafa}'' does not appear to be a \\commonly known entity. \\User: \{Generate prompt illustrated in Table \ref{prompt_en}\}\\Response: \textbf{Sakorafa} has made significant contributions to the \\field of environmental conservation and sustainable agriculture.\end{tabular}  \\
\bottomrule
\end{tabular}
}
\label{case}
\vspace{-0.5cm}
\end{table}

\subsection{The Generalization of HeardU}
\paragraph{LLM} Table \ref{case} shows several cases generated by the LLM. Note that in the first case, the LLM can generate realistic and fluent sentences using its inherent knowledge of a given entity and type. 
For the second one, we first ask whether the LLM understands the entity, and then ask the LLM to generate the corresponding sentence given the entity and entity type. The LLM is capable of producing a relevant sentence successfully, as it has encountered a sufficient number of entities in its training data. Even in cases where the LLM has not previously encountered the specific entity, it can generate plausible sentences by inferring meaning based on the entity's name and type.
\paragraph{TTS} The TTS model is trained on tens of millions of hours of speech data. For the En, its pronunciation and spelling are strongly related, allowing for the generation of relatively accurate speech from given text. This enables synthesis of the speech of an entity that has never been seen before. Conversely, the structure of the Zh relies on a multitude of characters, most of which can be correctly generated. For extremely rare characters, the TTS system employs a mapping table to convert them into simpler characters with the same pronunciation. In short, the TTS model demonstrates strong generalization capabilities, allowing it to adapt to a wide array of linguistic scenarios.

\subsection{The Impact of Noise Data}
\begin{figure}[t!]
\centering
\includegraphics[width=0.7\linewidth]{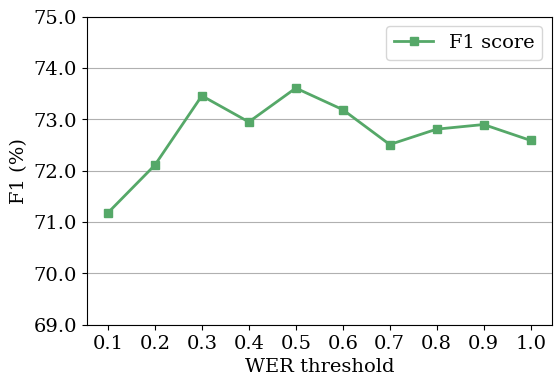}
    \vspace{-0.3cm}
  \caption {The performance of our method in the in-domain En setting with different WER thresholds.}
  \label{fig: wer}
\vspace{-0.4cm}
\end{figure}


We explore the influence of thresholds using WER as the noise metric. As shown in Figure \ref{fig: wer}, our model performs best with an F1 score of 73.78\% when using a WER threshold of 0.5.
Note that when the threshold is high, there is too much noisy data in the pseudo-data, which degrades the model's performance.
Conversely, when the threshold is set too low ($<$0.2), most of the noisy data is filtered out, resulting in overly clean training data that makes the model less robust to the noisy data in the test set.
Therefore, taking a relatively intermediate threshold can not only eliminate the excessively noisy part of the pseudo-data, but also retain some noisy data to enhance the robustness of the model.


\subsection{The Impact of Entity Frequency}

We analyze the influence of different data sizes, which represent the proportion of entities used. As shown in Table \ref{data_size}, with the decrease in data size, the overall WER does not increase much, but F1 declines greatly. 
This indicates that reducing the number of instances for each entity has a great impact on the final NER performance.
\begin{table}[t!]
\caption{Influence of different data sizes in the in-domain En setting.}
\renewcommand\arraystretch{1.1}
\resizebox{\columnwidth}{!}{
\centering
\begin{tabular}{lccccc} 
\hline
Data Size & WER$\downarrow$         & Precision$\uparrow$     & Recall$\uparrow$        & F1$\uparrow$            & \multicolumn{1}{l}{Label-F1$\uparrow$}  \\ 
\hline
100\%     & \textbf{\textbf{14.45}} & \textbf{\textbf{77.53}} & \textbf{\textbf{70.37}} & \textbf{\textbf{73.78}} & 
79.12                 \\
75\%      & 14.58                   & 77.20                   & 69.25                   & 73.01                   & \textbf{\textbf{79.30}}                                   \\
50\%      & 14.70                   & 76.05                   & 68.03                   & 71.82                   & 78.44                                   \\
25\%      & 15.15                   & 76.96                   & 63.69                   & 69.70                   & 76.12                                   \\
\hline
\end{tabular}
}
\label{data_size}
\vspace{-0.5cm}
\end{table}




\section{Conclusion}
It is challenging even for humans to identify and categorize an unfamiliar entity that they have never heard of.
In this paper, we show that constructing and maintaining an NED is feasible. We can present entities in the NED to Spoken NER models by generating synthetic data with LLMs and TTS models.
We further analyze the noise in the synthetic data and propose a noise metric for data filtering.
We achieve significant improvement across different settings, confirming the flexibility and generalization of the proposed HeardU framework.
We hope the released resources will promote further development in this area.

\section*{Acknowledgment}
The project was supported by National Natural Science Foundation of China (No. 62036004, No. 62276219), Natural Science Foundation of Fujian Province of China (No. 2024J011001), and the Public Technology Service Platform Project of Xiamen (No.3502Z20231043). We also thank the reviewers for their insightful comments.

\clearpage


\bibliographystyle{IEEEbib}
\bibliography{ref}

\vspace{12pt}

\end{document}